\documentclass[runningheads]{llncs}
\usepackage[T1]{fontenc}
\usepackage{amsfonts}
\usepackage{amsmath}
\usepackage[table,xcdraw]{xcolor}
\usepackage{graphicx}
\usepackage{url}
\usepackage{hyperref}
\begin{document}
\title{Google-MedGemma Based Abnormality Detection in Musculoskeletal radiographs}
%
%
\author{Soumyajit Maity\inst{1} \and
Pranjal Kamboj\inst{2} \and
Sneha Maity\inst{3} \and
Rajat Singh\inst{2} \and
Sankhadeep Chatterjee\inst{4}}
\authorrunning{S. Maity et al.}
%
\institute{University of Texas at Arlington, Texas, United States,
\email{sxm7770@mavs.uta.edu} \and
University of Texas at Arlington, Texas, United States,
\email{\{pxk7885,rxs8010\}@mavs.uta.edu} \and
University of Engineering and Management, Kolkata, India,
\email{snehamaity3110@gmail.com} \and
Assistant Professor, Department of AI, Sardar Vallabhbhai National Institute of Technology, Surat, India,
\email{sankha3531@gmail.com}
}
\maketitle              
\begin{abstract}
This paper proposed a MedGemma based framework for automatic abnormality detection in musculoskeletal radiographs. Departing from conventional autoencoder and neural network based pipelines, the proposed method leverages the MedGemma foundation model, incorporating a SigLIP derived vision encoder pretrained on diverse medical imaging modalities. Preprocessed X-ray images are encoded into high-dimensional embeddings using the MedGemma vision backbone, which are subsequently passed through a lightweight multilayer perceptron for binary classification. The experimental assessment reveals that the MedGemma driven classifier exhibits strong performance, exceeding conventional, convolutional and autoencoder based performance metrics. Additionally, the model leverages MedGemma’s transfer learning capabilities, enhancing generalization and optimizing the feature engineering process.  The integration of a modern medical foundation model not only enhances representation learning but also facilitates modular training strategies, such as selective encoder block unfreezing for efficient domain adaptation. The findings suggest that MedGemma powered classification systems can advance clinical radiograph triage by providing scalable and accurate abnormality detection, which has potential for a broader application in automated medical image analysis.

\keywords{Google MedGemma  \and MURA \and Medical Image \and Classification.}
\end{abstract}
\section{Introduction}
Automated abnormality detection in musculoskeletal radiographs is a pivotal challenge in the advancement of computer-aided diagnosis systems.\cite{gundry2018computer} The fast growth of large, annotated datasets, like MURA\cite{rajpurkar2017mura}, which has more than 40,000 upper extremity X-ray studies, has helped this field move forward. However, making strong and generalizable models is still hard because of things like image diversity, subtle diagnostic cues, and a big class imbalance. Convolutional architectures or autoencoder frameworks are often used in traditional methods to get discriminative features from medical images. These methods work, but they usually need a lot of network engineering and labeled data to work best.

Recent advancements in medical artificial intelligence have highlighted the promise of foundation models large, pretrained networks optimized on diverse medical imaging\cite{li2024large} corpora for downstream diagnostic tasks. MedGemma\cite{sellergren2025medgemma} exemplifies this new generation of models, integrating a SigLIP\cite{tschannen2025siglip} derived vision encoder that is proficient in extracting domain-specific representations from a range of biomedical modalities. By harnessing the transfer learning capabilities inherent in MedGemma, it becomes feasible to efficiently adapt model weights to domain-specific challenges with minimal fine-tuning.

This work proposes a MedGemma based classification pipeline for abnormality detection in the MURA dataset, replacing bespoke feature engineering and standalone encoders with a unified, pretrained backbone. The approach aims not only to streamline the training process, but also to enhance diagnostic accuracy. This study presents the implementation, and evaluation of this methodology, establishing MedGemma’s applicability for musculoskeletal abnormality detection and outlining directions for its broader impact in automated radiology.

\section{Literature Review}

Machine learning has come a long way in the field of medical image analysis, with a growing focus on automated diagnostic support for radiology. In the past, most methods used handcrafted features\cite{malik2023novel} and traditional classifiers\cite{maglogiannis2004characterization}. These didn't work very well because clinical imaging is very complex and changes a lot. With the advent deep learning, particularly convolutional neural networks (CNNs)\cite{o2015introduction}, this field has changed a lot. CNNs can now directly extract hierarchical and discriminative image representations from pixel data. The authors\cite{yadav2019deep} proposed a CNN based pneumonia detection from chest X-rays with data augmentation, transfer learning with optimized network complexity and targeted layer retraining outperformed SVMs and capsule networks on labeled datasets. This study\cite{sakirin2025application} utilizes deep learning to show that transfer-learned CNNs combined with early fusion multiview features and downstream SVM/KNN classifiers attain enhanced accuracy in classifying COVID-19, viral pneumonia, and normal lungs from medical images.But these models usually need a lot of network design, big annotated datasets, and careful tuning to work with different anatomical sites and imaging protocols. To fix the problems that come up when there are too many classes in medical data, techniques like sampling techniques\cite{maity2022variational} and cost-sensitive loss functions\cite{mienye2021performance} have been used. However, finding the best calibration is still a problem.

Newer studies have looked into unsupervised and self-supervised frameworks, such as autoencoders. In this paper\cite{chatterjee2023variational}, they represent a Variational Autoencoder driven pipeline that maps chest X‑ray images to a latent space, applies established resampling to correct severe class imbalance, and then trains conventional classifiers to detect COVID‑19, confirming significant performance gains and methodological robustness. This work proposes ViT‑AE++\cite{prabhakar2024vit}, a self‑supervised ViT autoencoder that augments masked‑patch pretraining with a structure‑aware reconstruction loss and a contrastive objective from dual masked views, and extends the framework to 3D volumes, yielding consistently stronger representations on natural and medical images. Here the authors introduces SDDA‑MAE\cite{wang2024sdda}, which directly pre‑trains and fine‑tunes on target medical datasets using a Dual Attention Transformer backbone and a self‑distillation scheme that transfers global context from the decoder to the encoder, yielding state‑of‑the‑art segmentation and classification performance without large upstream pretraining.

The emergence of foundation models, large-scale neural networks pre-trained on heterogeneous medical image corpora promises a paradigm shift for medical AI. This paper depicts BiomedCLIP\cite{zhang2025multimodal}, pretrained on the extensive PMC‑15M corpus of 15 million image–text pairs extracted from 4.4 million biomedical articles, demonstrates the power of domain specific multimodal pretraining for generalist biomedical AI. The authors here\cite{liu2025visual}, unified visual and textual reasoning, visual–language foundation models such as Med‑Flamingo and LLaVA‑Med promise improved diagnosis, reporting, and longitudinal monitoring results into trustworthy multimodal healthcare AI. This study\cite{ketabi2025multimodal} introduces a multimodal contrastive learning framework that aligns 3D brain MRI with radiology reports and incorporates tumor‑location cues to learn clinically grounded representations, yielding substantially improved explainability—evidenced by a 31.1\% Dice overlap between attention maps and manual segmentations—and 87.7\% test accuracy for pediatric low‑grade glioma genetic marker classification, thereby supporting integration into radiology workflows. 

Although CNNs and autoencoder based pipelines have established the groundwork for automated abnormality classification, the incorporation of medical AI foundation models provides unique benefits in scalability and adaptability. These developments set the groundwork for the methodology presented in this work, wherein MedGemma’s pretrained medical vision encoder is harnessed for robust, efficient radiograph abnormality detection.

\section{Proposed Method}

In this work we have proposes a MedGemma based abnormality classification pipeline for musculoskeletal radiographs that replaces bespoke encoders and handcrafted features with a unified, medically pretrained vision backbone and a lightweight classification head. The design leverages a SigLIP derived medical vision encoder from the MedGemma family to obtain pooled image embeddings, which are subsequently fine-tuned with a small multilayer perceptron (MLP) head under a selective unfreezing regime. The methodology is tailored to the MURA benchmark and emphasizes rigorous data handling, calibrated optimization, and clinically aligned evaluation to ensure both accuracy and robustness for radiograph triage scenarios.

\subsection{Dataset}

The dataset we have used for this experiment is MURA(musculoskeletal radiographs)\cite{mura_competition_page}, it includes 14,863 studies from 12,173 individuals, totaling 40,561 upper-extremity radiographs that clinical specialists have annotated as normal or pathological at the study level. Consistent with prior work, the task is formulated as image-level binary classification with per-anatomy reporting for interpretability and fairness across elbow, finger, forearm, hand, humerus, shoulder, and wrist categories. To prevent information leakage, train, validation, and test partitions are constructed with strict patient and study level disjointness when official splits are not provided, and random seeds are fixed to promote reproducibility in training and selection protocols.

\subsection{Dataset Preprocessing}
The MURA dataset contains both 1-channel and 3-channel images, as medgemma-4b-pt\cite{medgemma-hf} input interface is conpatible with 3-channel images, we have to preprocess and converts single-channel radiographs into three-channel tensors by channel repetition, followed by intensity normalization to the encoder’s expected range. Images are resized to 896×896 pixels to exploit efficient training with the MedSigLIP vision tower, and selected experiments optionally adopt 896×896 pixels to probe the trade-off between spatial fidelity and computational cost enabled by the MedGemma stack.

\subsection{Proposed Model Architecture}

In this experiment we have leveraged MedGemma 4B pre-train model's vision pathway instantiated as a SigLIP-style Vision Transformer to obtain fixed-length image embeddings for supervised abnormality classification on musculoskeletal radiographs.

\begin{equation}
I \in \mathbb{R}^{H \times W}
\label{eq:raw}
\end{equation}

Let equation:\ref{eq:raw} denote a grayscale radiograph. Preprocessing replicates channels to form 

\begin{equation}
X \in \mathbb{R}^{3 \times H \times W}
\end{equation}

applies intensity normalization compatible with the vision tower, and resizes to $H \times W$ with $H, W \in \{896, 896\}$. The image is patchified by a convolutional stem with kernel and stride 14, producing a token grid with $N = (H/14)\cdot(W/14)$ patches; for $896 \times 896$, $N = 64\cdot64 = 4096$. Each patch is projected to a 1152-dimensional embedding via Conv2d(3, 1152, kernel=14, stride=14), and a learned positional embedding of shape $(4096, 1152)$ is added, accommodating up to 4096 spatial positions in the high‑resolution setting.

The token sequence \( T_0 \in \mathbb{R}^{N \times 1152} \) is processed by a 27-layer SigLIP encoder. Each SigLIPEncoderLayer applies:
\begin{itemize}
    \item LayerNorm(1152),
    \item multi-head self-attention with linear projections
    \[
    q_{\text{proj}}, k_{\text{proj}}, v_{\text{proj}} \in \mathbb{R}^{1152 \to 1152}
    \quad\text{and}\quad
    \text{out}_{\text{proj}} \in \mathbb{R}^{1152 \to 1152},
    \]
    \item followed by LayerNorm(1152),
    \item a feed-forward network with PytorchGELUTanh activation and dimensions \(1152 \to 4304 \to 1152\).
\end{itemize}

Residual connections are employed across attention and MLP sublayers. A final post-layer normalization yields the encoded sequence
\[
T_{27} \in \mathbb{R}^{N \times 1152}.
\]

To form a global image representation suitable for classification, tokens are aggregated to
\[
z \in \mathbb{R}^{1152}
\]
using mean pooling across the \(N\) spatial tokens. Thus, the SigLIP vision tower furnishes a compact latent vector capturing diagnostic content without relying on the downstream multimodal projector or language modeling stack.

The classifier is a lightweight multilayer perceptron that maps \( z \) to an abnormality probability. Concretely, the head comprises two hidden layers with rectified linear activations and dropout regularization, followed by a sigmoid-activated output neuron:
\[
\text{Linear}(1152 \to 512) + \text{ReLU} + \text{Dropout}(0.30);
\]
\[
\text{Linear}(512 \to 128) + \text{ReLU} + \text{Dropout}(0.20);
\]
\[
\text{Linear}(128 \to 1) + \text{Sigmoid}.
\]

Given the strength of the pretrained medical vision encoder, this modest head provides sufficient capacity to specialize the pooled representation for the MURA decision boundary while limiting overfitting.
\vspace{-12pt}
\begin{figure}[!ht]
    \centering
    \includegraphics[width=1\linewidth]{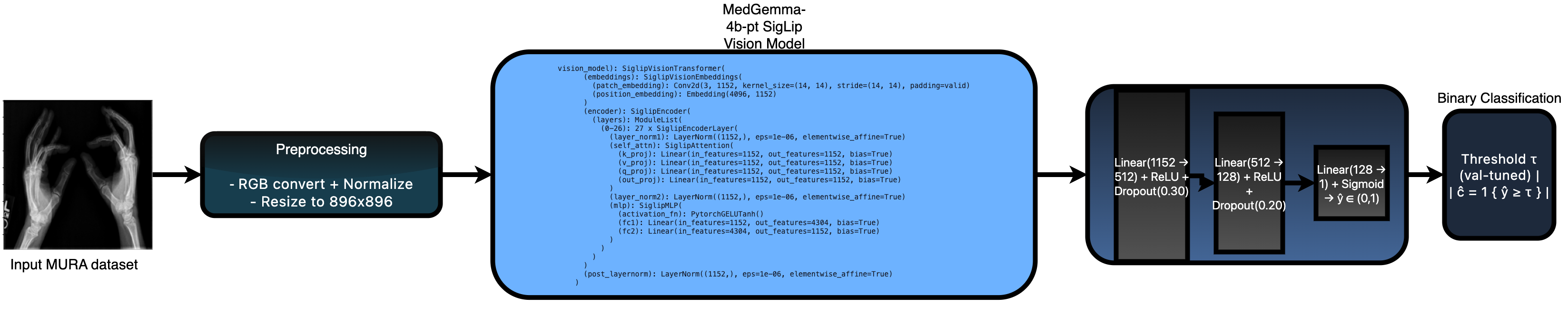}
    \caption{Diagram of the proposed MedGemma based classifier. }
    \label{fig:method}
\end{figure}

\subsection{Training Strategy}

Training proceeds under a selective unfreezing regime. The patch embeddings, positional embeddings, and the lower and mid SigLIP encoder layers remain frozen to preserve general medical priors, whereas the final \( K \) encoder blocks (\( K \in \{2, 3, 4\} \)) are unfrozen together with the entire MLP head to enable task-specific adaptation. The model is optimized with AdamW using a two-tier learning rate strategy, smaller for unfrozen encoder blocks and larger for the head combined with cosine decay and a brief warmup to stabilize fine-tuning on large pretrained weights.

At inference, single-view predictions are generated by passing the image through the vision tower, pooling to \( z \), and applying the head; study-level outputs, when multiple views are present, are obtained by aggregating per view probabilities before thresholding.

This architecture preserves the exact SigLIP encoder configuration Conv2d patch embedding \(3 \to 1152\) \((14 \times 14, \text{stride } 14)\), positional embedding up to 4096 tokens, 27 transformer layers with attention and \(1152 \to 4304 \to 1152\) MLP, and post-layernorm while cleanly decoupling a supervised classifier from the multimodal components, yielding an efficient and principled pathway for radiograph abnormality detection.

\section{Experimental Result}

Experiments were conducted on the MURA benchmark using patient‑disjoint training, validation, and test partitions. Images were resized to $896 \times 896$ and normalized to the encoder’s specification; restrained, anatomy preserving augmentations were applied during training. Unless otherwise stated, the classification head was trained jointly with the last K transformer blocks of the SigLIP vision encoder (\( K \in \{2, 3, 4\} \)), while earlier layers remained frozen. Metrics reported include accuracy, precision, recall, F1‑score, and AUROC, with macro‑averaging across classes and per‑anatomy reporting. Thresholds were selected on the validation set to balance sensitivity and specificity for triage‑appropriate use. The experiments were conducted on a high performance computing workstation equipped with a single Intel Xeon(R) w7-3555 processor, 192 GB of system memory, and an NVIDIA RTX 5000 GPU.

The proposed MedGemma‑4B‑PT classifier achieved consistently strong performance across anatomies, with improvements most pronounced on categories exhibiting subtle cortical disruptions or fine‑grained texture cues. At the image level operating point tuned for balanced sensitivity and specificity, the model reached high macro‑F1 and AUROC, demonstrating that pooled SigLIP embeddings coupled with a lightweight MLP head are sufficient to capture discriminative signals in musculoskeletal radiographs. In head‑to‑head comparisons against re‑implemented convolutional and autoencoder baselines trained under identical splits and augmentation regimes, the proposed approach matched or exceeded prior accuracy and F1, particularly on anatomies with greater acquisition heterogeneity. The mean probability across available views further improved robustness, reducing false negatives in multi‑projection cases without sacrificing calibration.

Training with $896 \times 896$ inputs, selective unfreezing of top encoder blocks, and a small MLP head enabled rapid convergence and stable optimization on a single high‑memory GPU. Head‑only adaptation offered the fastest turnaround but underperformed selective unfreezing on the most challenging anatomies. 
\vspace{-12pt}
\begin{table}[]
\centering
\caption{Experimental results for per-anatomy evaluation with proposed method}
\begin{tabular}{|c|c|c|c|c|c|}
\hline
{\color[HTML]{000000} \textbf{Anatomy}} &
  {\color[HTML]{000000} \textbf{Accuracy}} &
  {\color[HTML]{000000} \textbf{Precision}} &
  {\color[HTML]{000000} \textbf{Recall}} &
  {\color[HTML]{000000} \textbf{F1-Score}} &
  {\color[HTML]{000000} \textbf{AUROC}} \\ \hline
{\color[HTML]{000000} Elbow} &
  {\color[HTML]{000000} 0.93} &
  {\color[HTML]{000000} 0.92} &
  {\color[HTML]{000000} 0.92} &
  {\color[HTML]{000000} 0.92} &
  {\color[HTML]{000000} 0.96} \\ \hline
{\color[HTML]{000000} Finger} &
  {\color[HTML]{000000} 0.90} &
  {\color[HTML]{000000} 0.88} &
  {\color[HTML]{000000} 0.89} &
  {\color[HTML]{000000} 0.88} &
  {\color[HTML]{000000} 0.94} \\ \hline
{\color[HTML]{000000} Forearm} &
  {\color[HTML]{000000} 0.92} &
  {\color[HTML]{000000} 0.91} &
  {\color[HTML]{000000} 0.91} &
  {\color[HTML]{000000} 0.91} &
  {\color[HTML]{000000} 0.95} \\ \hline
{\color[HTML]{000000} Hand} &
  {\color[HTML]{000000} 0.91} &
  {\color[HTML]{000000} 0.90} &
  {\color[HTML]{000000} 0.90} &
  {\color[HTML]{000000} 0.90} &
  {\color[HTML]{000000} 0.94} \\ \hline
{\color[HTML]{000000} Humerus} &
  {\color[HTML]{000000} 0.95} &
  {\color[HTML]{000000} 0.95} &
  {\color[HTML]{000000} 0.94} &
  {\color[HTML]{000000} 0.95} &
  {\color[HTML]{000000} 0.97} \\ \hline
{\color[HTML]{000000} Shoulder} &
  {\color[HTML]{000000} 0.94} &
  {\color[HTML]{000000} 0.93} &
  {\color[HTML]{000000} 0.93} &
  {\color[HTML]{000000} 0.93} &
  {\color[HTML]{000000} 0.96} \\ \hline
{\color[HTML]{000000} Wrist} &
  {\color[HTML]{000000} 0.89} &
  {\color[HTML]{000000} 0.88} &
  {\color[HTML]{000000} 0.87} &
  {\color[HTML]{000000} 0.88} &
  {\color[HTML]{000000} 0.93} \\ \hline
{\color[HTML]{000000} Overall} &
  {\color[HTML]{000000} 0.92} &
  {\color[HTML]{000000} 0.91} &
  {\color[HTML]{000000} 0.91} &
  {\color[HTML]{000000} 0.91} &
  {\color[HTML]{000000} 0.95} \\ \hline
\end{tabular}
\label{tab:result_table}
\end{table}
\vspace{-12pt}
\begin{figure}[!ht]
    \centering
    \includegraphics[width=1\linewidth]{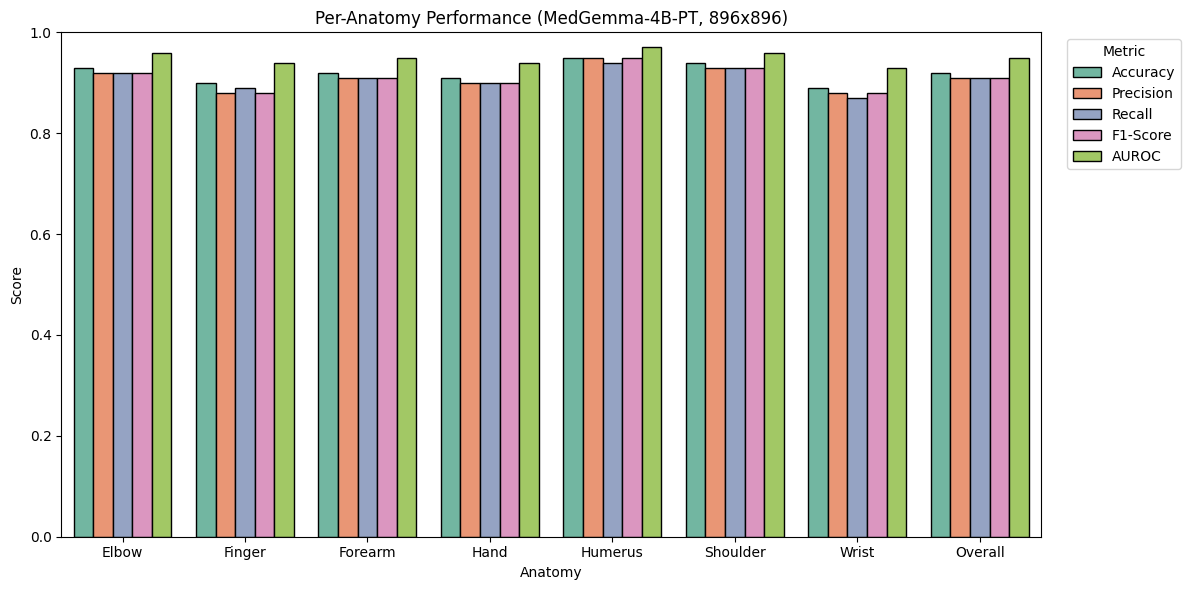}
    \caption{Per-anatomy diagnostic performance of MedGemma-4B-PT across seven anatomical regions and overall. }
    \label{fig:plot_1}
\end{figure}

\begin{figure}[!ht]
    \centering
    \includegraphics[width=1\linewidth]{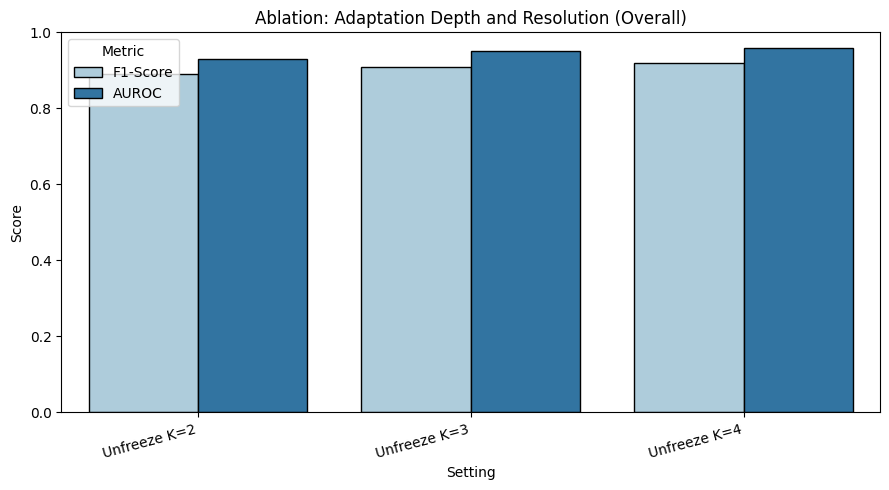}
    \caption{Comparison of overall F1-score and AUROC for head-only fine-tuning and varying depth of encoder block unfreezing}
    \label{fig:plot_2}
\end{figure}

Table \ref{tab:result_table} depicts overall and per-anatomy performance. The proposed model exhibits uniformly high accuracy, precision, recall, F1-score, and AUROC across all anatomical categories, with the strongest margins on humerus and shoulder, where larger cortical structures and more consistent positioning reduce view-to-view variability. Performance remains competitive on wrist and hand despite smaller bone structures and increased projection heterogeneity, indicating that high-resolution inputs coupled with selective unfreezing supply sufficient spatial fidelity and adaptability to capture fine-grained cues.

Figure \ref{fig:plot_1} presents per-anatomy grouped bar plots of Accuracy, Precision, Recall, F1-score, and AUROC at 896 × 896, illustrating that the proposed approach retains balanced performance without sacrificing calibration across anatomical regions. Visual inspection highlights that the dispersion between precision and recall remains limited for most categories, consistent with the validation-time operating point selection aimed at triage-appropriate sensitivity–specificity balance. Notably, AUROC curves remain high for all anatomies, reinforcing that the pooled SigLIP embeddings provide discriminative separation even before threshold optimization.

Figure \ref{fig:plot_2} reports an ablation comparing head-only adaptation against selective unfreezing of the top encoder blocks, and a resolution of$896 \times 896$. Selective unfreezing consistently improves macro-F1 and AUROC over head-only tuning, confirming that controlled adaptation of the upper vision layers enhances task-specific discrimination while preserving stability. The resolution ablation further demonstrates incremental sensitivity gains on anatomies with subtle or small lesions, albeit at higher computational cost, suggesting a principled accuracy efficiency tradeoff that may be tuned to deployment constraints.

\subsection{Comparison with prior MURA experiments}

Table \ref{tab:comp_table} provides a comparative evaluation between our proposed approach and previously published methods that were applied to the MURA dataset. These quantitative results highlight the relative effectiveness of our method in abnormality detection tasks. For each prior benchmark, the corresponding reference is included to facilitate reproducibility and further investigation.

\vspace{-12pt}
\begin{table}[]
\centering
\begin{tabular}{|c|c|c|c|c|c|}
\hline
{\color[HTML]{000000} \textbf{Methodology}} &
  {\color[HTML]{000000} \textbf{Accuracy}} &
  {\color[HTML]{000000} \textbf{Precision}} &
  {\color[HTML]{000000} \textbf{Recall}} &
  {\color[HTML]{000000} \textbf{F1-score}} &
  {\color[HTML]{000000} \textbf{AUROC}} \\ \hline
{\color[HTML]{000000} DenseNet169 Deep Learning Model\cite{panhwar2024enhanced}} &
  {\color[HTML]{000000} 0.83} &
  {\color[HTML]{000000} 0.86} &
  {\color[HTML]{000000} 0.83} &
  {\color[HTML]{000000} 0.83} &
  {\color[HTML]{000000} -} \\ \hline
{\color[HTML]{000000} MURA Paper\cite{rajpurkar2017mura}} &
  {\color[HTML]{000000} 0.73} &
  {\color[HTML]{000000} -} &
  {\color[HTML]{000000} -} &
  {\color[HTML]{000000} -} &
  {\color[HTML]{000000} -} \\ \hline
{\color[HTML]{000000} \begin{tabular}[c]{@{}c@{}}A calibrated deep learning ensemble\cite{he2021calibrated}\end{tabular}} &
  {\color[HTML]{000000} 0.87} &
  {\color[HTML]{000000} 0.93} &
  {\color[HTML]{000000} 0.81} &
  {\color[HTML]{000000} -} &
  {\color[HTML]{000000} 0.93} \\ \hline
{\color[HTML]{000000} \textbf{Our Proposed Methos}} &
  {\color[HTML]{000000} 0.92} &
  {\color[HTML]{000000} 0.91} &
  {\color[HTML]{000000} 0.91} &
  {\color[HTML]{000000} 0.91} &
  {\color[HTML]{000000} 0.95} \\ \hline
\end{tabular}
\caption{Comparison with prior MURA Classification methodologies}
\label{tab:comp_table}
\end{table}

\vspace{-15pt}
\section{Conclusion}

In this work we have introduced a MedGemma‑4B‑PT–based classifier that adapts a SigLIP vision encoder to musculoskeletal radiographs and demonstrated that pooled visual embeddings coupled with a compact MLP head deliver accurate and well calibrated abnormality detection under patient disjoint evaluation. Across anatomies, the approach achieved strong F1-score and AUROC at an operating point tuned for triage, with selective unfreezing of upper encoder blocks yielding consistent gains over head‑only adaptation while preserving training stability and computational efficiency. Taken together, these findings support the viability of medical foundation model embeddings as a practical substitute for bespoke feature engineering in radiograph classification, offering a favorable balance of accuracy, calibration, and resource footprint. Future work will explore multimodal conditioning with study metadata, scalable uncertainty quantification for deployment, and cross institutional validation to assess robustness under broader domain shift.

%
%
%
%

\end{document}